\documentclass[sigconf]{acmart}

\AtBeginDocument{%
 \providecommand\BibTeX{{%
 \normalfont B\kern-0.5em{\scshape i\kern-0.25em b}\kern-0.8em\TeX}}}

\usepackage{amsfonts}
\usepackage{bbm}
\usepackage{algorithm}
\usepackage{algorithmic}
\usepackage{array}
\usepackage{xcolor}
\usepackage{multirow}
\usepackage{stfloats}
\usepackage{verbatim}
\usepackage{graphicx}
\usepackage{booktabs}
\usepackage[precent]{overpic}

\definecolor{forestgreen}{HTML}{009B55} 

\definecolor{commentcolor}{RGB}{110,154,155} 

\AtBeginDocument{%
  \providecommand\BibTeX{{%
    Bib\TeX}}}

\setcopyright{acmcopyright}
\copyrightyear{2023}
\acmYear{2023}
\setcopyright{acmlicensed}
\acmConference[MM '23] {Proceedings of the 31st ACM International Conference on Multimedia}{October 29--November 3, 2023}{Ottawa, ON, Canada.}
\acmBooktitle{Proceedings of the 31st ACM International Conference on Multimedia (MM '23), October 29--November 3, 2023, Ottawa, ON, Canada}
\acmPrice{15.00}
\acmISBN{979-8-4007-0108-5/23/10}
\acmDOI{10.1145/3581783.3611794}

\acmSubmissionID{428}

\begin{document}
\begin{sloppypar}
\title{Disentangling Multi-view Representations Beyond Inductive Bias}


\author{Guanzhou Ke}
\orcid{0000-0002-1812-367X}
\email{guanzhouk@gmail.com}
\affiliation{%
  \institution{Beijing Jiaotong Univeristy}
  \city{Beijing}
  \country{China}
}

\author{Yang Yu}
\authornote{Corresponding authors.}
\orcid{0000-0003-1265-0433}
\email{yangy1@bjtu.edu.cn}
\affiliation{
  \institution{Beijing Jiaotong Univeristy}
  \city{Beijing}
  \country{China}
}

\author{Guoqing Chao}
\orcid{0000-0002-2410-650X}
\email{guoqingchao@hit.edu.cn}
\affiliation{
  \institution{Harbin Institute of Technology}
  \city{Weihai}
  \country{China}
}

\author{Xiaoli Wang}
\orcid{0000-0001-9336-1013}
\email{xiaoliwang@njust.edu.cn}
\affiliation{
  \institution{Nanjing University of Science and Technology}
  \city{Nanjing}
  \country{China}
}

\author{Chenyang Xu}
\orcid{0000-0003-1174-7233}
\email{chyond.xu@gmail.com}
\affiliation{%
  \institution{Wuyi University}
  \city{Jiangmen}
  \country{China}
}

\author{Shengfeng He}
\authornotemark[1]
\orcid{0000-0002-3802-4644}
\email{shengfenghe@smu.edu.sg}
\affiliation{%
  \institution{Singapore Management University}
  \country{Singapore}
}

\renewcommand{\shortauthors}{Ke et al.}

\begin{abstract}
  Multi-view (or -modality) representation learning aims to understand the relationships between different view representations. Existing methods disentangle multi-view representations into consistent and view-specific representations by introducing strong inductive biases, which can limit their generalization ability. In this paper, we propose a novel multi-view representation disentangling method that aims to go beyond inductive biases, ensuring both interpretability and generalizability of the resulting representations. Our method is based on the observation that discovering multi-view consistency in advance can determine the disentangling information boundary, leading to a decoupled learning objective. We also found that the consistency can be easily extracted by maximizing the transformation invariance and clustering consistency between views. These observations drive us to propose a two-stage framework. In the first stage, we obtain multi-view consistency by training a consistent encoder to produce semantically-consistent representations across views as well as their corresponding pseudo-labels. In the second stage, we disentangle specificity from comprehensive representations by minimizing the upper bound of mutual information between consistent and comprehensive representations. Finally, we reconstruct the original data by concatenating pseudo-labels and view-specific representations. Our experiments on four multi-view datasets demonstrate that our proposed method outperforms 12 comparison methods in terms of clustering and classification performance. The visualization results also show that the extracted consistency and specificity are compact and interpretable. Our code can be found at \url{https://github.com/Guanzhou-Ke/DMRIB}.
\end{abstract}




\maketitle

\section{Introduction}


\begin{figure}[t!]
\centering
\includegraphics[width=\linewidth]{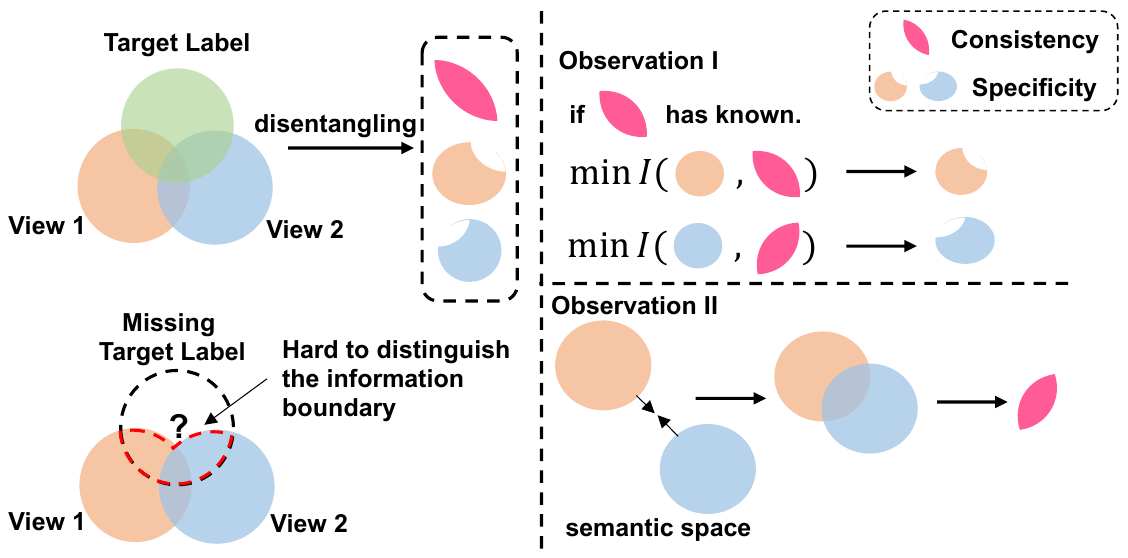}
\caption{Illustration of the key observations and basic idea behind our method. On the left side of the graph, when target labels are available, we can disentangle consistent (pink) and specific representations (orange and blue) from comprehensive representations using the information bottleneck principle. However, when target labels (e.g., object categories) are missing, it becomes difficult to distinguish the boundaries of each part of the information. Based on this phenomenon, we found (on the right side) that if multi-view consistency is known in advance, we can obtain the remaining information by minimizing the mutual information between the comprehensive and consistent representations. Additionally, we observed that pulling the semantic distance between two views closer in a suitable semantic space helps us extract the multi-view consistency.}
\label{pic:observation}
\end{figure}

With the increasing availability of data from various sources, such as images, text, and sensors, it is essential to extract useful and rich information from them in multimedia applications. Multi-view Representation Learning (MRL)~\cite{wang2015deep}, also known as multi-modal representation learning, is a promising approach that has gained attention in the communities. In cross-modal retrieval~\cite{rasiwasia2010new}, ``views'' can be the pairs of images and corresponding textual descriptions referring to the same object. In autonomous driving~\cite{zhou2020end-3d}, ``views'' can also refer to video frames of the same object captured by cameras at different positions. While MRL has proven effective in practical applications, such as clustering~\cite{huang2023fast, fang2023efficient}, classification~\cite{9939043, wang2022mmatch}, and face synthesis~\cite{Xu_Li_Xu_He_2020, xu2022fully, xu2021multi-face}, understanding the underlying relationships between different view representations remains an open question. In general, multi-view comprehensive representations consist of consistent and specific representations~\cite{liu2014partially}, which combine in a certain pattern to form the multi-view comprehensive representation. Consistent representations refer to the shared information among different views, while view-specific representations refer to the private information of each view. Therefore, distinguishing between these two representations is an essential step in understanding multi-view representations.

One possible approach to understanding multi-view representations is to disentangle the comprehensive representation of each view and extract its consistency and specificity. Previous methods~\cite{DBLP:conf/iclr/Federici0FKA20, lin2022dual, wang2019deep, wan2021multi} utilize the information bottleneck principle~\cite{tishby99information} to achieve this goal. The information bottleneck principle is a supervised method that aims to minimize the mutual information between the input data $\bold{x}$ and its embedding representation $\bold{z}$, while simultaneously maximizing the mutual information between $\bold{z}$ and target label $\bold{y}$. In the context of multi-view representation learning, this principle can be used to extract consistent representations by maximizing the mutual information between views, and view-specific representations can be similarly obtained by minimizing it. However, since most multi-view data is large-scale and unlabeled, these methods cannot be directly applied in unsupervised settings. In unsupervised settings, models cannot directly determine which parts of the representation correspond to consistency or specificity, which may lead to suboptimal solutions (see left part of Figure~\ref{pic:observation}). For example, in the scenario of identifying clothes from different angles, we hope the model focuses on the inherent traits of the clothes (consistency), rather than who is wearing them (specificity). This task is easy to solve in the supervised setting but is challenging in the unsupervised setting where the model needs to distinguish between the boundaries of people and clothes. To address this issue, some works~\cite{gonzalez2018image, xu2021multi, gabbay2021scaling} introduce prior distribution assumptions for the process of disentangling consistent and view-specific representations. For example, ~\cite{xu2021multi} assumes that the consistency obeys the Gumbel distribution, and the specificity obeys the mixed Gaussian distribution. We argue that both the information bottleneck principle and prior distribution assumptions are strong ``inductive biases''. While the former requires data to have distinguishable information boundaries, the latter requires data to obey the prior distribution assumptions. However, these inductive biases may limit the scalability of downstream applications.

The main objective of this research is to investigate whether unsupervised multi-view representation disentangling can be achieved with weak inductive biases. This study is motivated by two key observations, which are illustrated in Figure~\ref{pic:observation}. The first observation is that it is easier to separate specificity from comprehensive representations when consistent information is known beforehand. The second observation is that finding a suitable semantic space and narrowing the semantic distance between views in this space can help extract view consistency. Therefore, the challenge of extracting consistency is transformed into finding a suitable transformation space that satisfies two characteristics: low-level and high-level information. From the low-level perspective, the semantics associated with different view data of the same object should remain the same after data augmentation. For example, the semantics associated with the term ``dog'' remains the same even after color transformation is applied to views of dogs with different backgrounds. This is called transformation invariance. From the high-level perspective, all views of similar objects should have the same clustering prototype. We refer this to as clustering consistency. By narrowing the semantic distance of multi-view data in the space with transformation invariance and clustering consistency, multi-view consistency can be extracted. Based on these insights, a two-stage multi-view representation disentanglement method is proposed in this paper.


In the first stage, we employ a consistent encoder that maximizes both the transformation invariance and clustering consistency to output the consistent representation and corresponding clustering pseudo-labels. To maximize intra-view consistency, we first apply data augmentation to each view to generate new data that preserves the original semantics. Then, we map these augmented views to the same semantic space and cluster them to assign the same pseudo-label to views with the same semantics. We also introduce a maximum entropy constraint to prevent assigning all instances to the same cluster. In the second stage, we use multiple view-specific encoders to extract comprehensive representations for each view. We then minimize the upper bound of mutual information between view-consistent representations and comprehensive representations of each view to obtain specificity for each view. We adopt the VAE architecture and concatenate the pseudo-labels and view-specific representations as input to the view-specific decoders to generate data. Our approach has two benefits: First, the two-stage architecture only requires variational inference for view-specific representations; and second, utilizing pseudo-labels to control the view-generation process can improve interpretability. In other words, the pseudo-labels output from the first stage provides interpretability for the consistent representation and can control the generation of view data corresponding to the class. In addition, we extract specificity with a probabilistic approach, using a mixture of Gaussian distributions to fit the specificity distribution, which enables the sampling of data with different styles. Combining the two attributes, we can ultimately achieve data generation with specified output classes and output styles. Extensive experimental results on four multi-view datasets demonstrate our superior performances against state-of-the-art methods. The main contributions of this paper can be summarized as follows:
\begin{itemize}
 \item We propose a two-stage unsupervised multi-view representation disentangling method that goes beyond inductive bias, requiring only the information of consistency to achieve disentanglement.
 \item We delve into multi-view consistency by mining view transformation invariance and clustering consistency. Ablation studies show that the quality of consistent representations directly affects the expressive ability of the model and the quality of disentanglement.
 \item Our proposed method outperforms state-of-the-art methods in terms of clustering and classification performance. Visualization results also demonstrate that the extracted consistency and specificity are compact and interpretable.
\end{itemize}

\section{Related Work}
\label{sec:related-works}
\subsection{Multi-view Representation Learning}
Multi-view representation learning \cite{wang2015deep} can be broadly categorized into two groups: statistic-based and deep learning-based methods. Statistic-based methods focus on extracting view-consistent representations using techniques such as CCA-based methods \cite{rasiwasia2010new, dhillon2011multi}, non-negative matrix factorization methods \cite{liu2013multi, wang2018multiview}, and subspace methods \cite{brbic2018multi, wang2019multi}. However, these methods have difficulty scaling up to high-dimensional and large-scale data scenarios \cite{rasiwasia2010new, dhillon2011multi}. Therefore, a large number of deep learning-based methods have been developed in recent years \cite{andrew2013deep, wang2015deep, trosten2021reconsidering, zhou2020end, zhang2019ae2, xu2021multi-face, xu2021multi, LU2023383}. A comprehensive review \cite{LiYZ19} provides an overview of the current development status of MRL. Our method belongs to the deep learning-based approaches. In unsupervised scenarios, most previous methods use generative models such as autoencoders \cite{andrew2013deep, wang2015deep, zhang2019ae2} or GANs \cite{zhou2020end} to extract multi-view comprehensive representations. However, there is a significant amount of redundant information in these representations that does not provide substantial help for downstream tasks. Therefore, many methods have begun to focus on the importance of disentangling multi-view representations, such as extracting view-consistent representations using contrastive learning \cite{DBLP:conf/icdm/KeZY22, tian2020contrastive, trosten2021reconsidering} or separating consistent and view-specific representations based on the theory of information bottleneck \cite{xu2021multi, DBLP:conf/iclr/Federici0FKA20}. Unfortunately, in unsupervised environments, the model has difficulty distinguishing the boundaries between different types of information. Unlike previous methods, we propose a two-stage disentanglement strategy that first extracts view-consistent representations using self-supervised learning and then uses them as known information to extract view-specific representations by minimizing the upper bound of mutual information between consistent representations and comprehensive representations.

\begin{figure*}[t]
\centering
\includegraphics[width=7in]{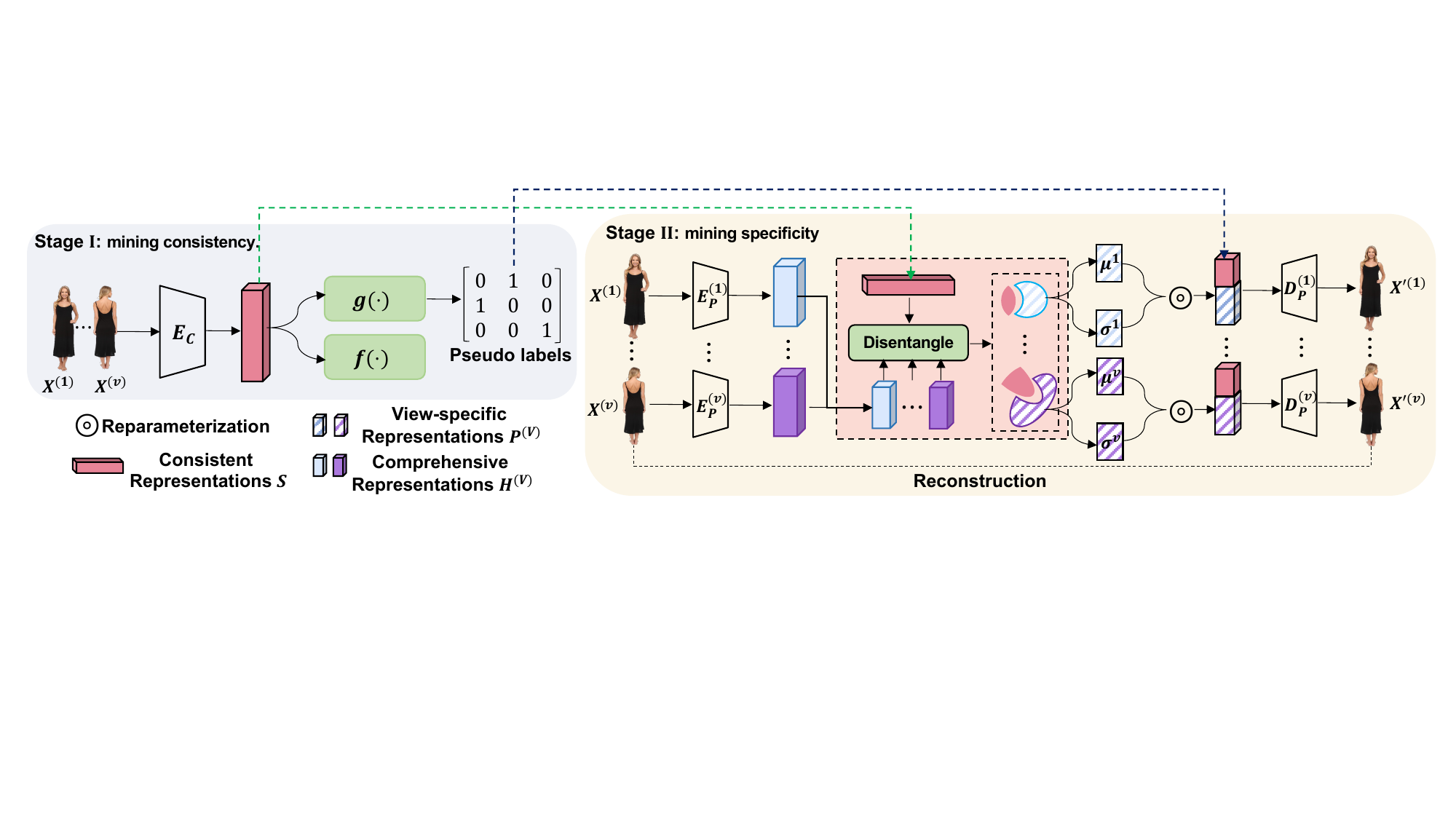}
\caption{Illustration of the workflow of the proposed method. It includes a consistent encoder $E_c(\cdot)$, $V$ view-specific encoders $E_p^{(i)}(\cdot)$ and decoders $D_p^{(i)}(\cdot)$, a clustering head $g(\cdot)$, a contrastive head $f(\cdot)$, and a disentanglement module. We adopt a two-stage approach to disentangle the consistency and the specificity. In the first stage, we train $E_c(\cdot)$ using contrastive and clustering heads, and then freeze its weights of the $E_c(\cdot)$. In the second stage, we use $E_p^{(i)}(\cdot)$ to learn the comprehensive representation of views, and then disentangle view-specific representations from the comprehensive representation by using the consistent representation $S$ and clustering labels as known information. Finally, we complete the reconstruction task by concatenating the view-specific representation and clustering labels as the input to $D_p^{(i)}(\cdot)$.}
\label{pic:framework}
\end{figure*}

\subsection{Disentangled Representation Learning}
Disentangled Representation Learning (DRL) has emerged as a promising direction for learning independent factors in representations. This concept can be traced back to Independent Component Analysis (ICA)\cite{yang1997adaptive}, which has since inspired various deep learning-based approaches, including InfoGAN\cite{chen2016infogan} and $\beta$-VAE~\cite{higgins2017beta}. VAE-based methods, in particular, learn a variational distribution $q(z) \sim \mathcal{N}(\mathbf{0}, \mathbf{1})$ to approximate the original data distribution $q(x)$, making them highly interpretable in statistics.

Recent efforts have focused on applying DRL to multi-view learning, where multiple views of data are available for training. For example, Xu et al.\cite{xu2021multi} proposed a multi-view DRL method that combines a generative model and a disentangled representation model to learn discriminative representations. Federici et al.\cite{DBLP:conf/iclr/Federici0FKA20} proposed a multi-view DRL method that applies a mutual information maximization objective to learn a joint latent space. Wang et al.~\cite{wang2019deep} proposed a deep multi-view learning method that incorporates a shared disentangled representation into the learning process.

In comparison to these existing methods, our proposed method only aims to separate the specificity from the comprehensive representations. This simplification allows us to use a VAE-based architecture for the second stage of our method. Additionally, we only need to perform variational inference on the specificity, which reduces the computational complexity of our approach. Furthermore, we use pseudo-labels to assist the data reconstruction process, which helps to generate the required data under certain conditions.


\section{Methodology}
\label{sec:Methodology}

Give a multi-view dataset with $V$ views $\mathcal{X}=\{ X^{(1)}, X^{(2)}, \cdots , X^{(V)} | X^{(i)} \in \mathbb{R}^{N \times d_v}\}$, where $d_v$ is the dimensionality of $v$-th view. The proposed method  aims to extract comprehensive representations from $\mathcal{X}$ and subsequently disentangle them into view-consistent representations $S$ and view-specific representations $P^{(v)}$. To this end,  we leverage unsupervised pre-text tasks, such as contrastive learning~\cite{chen2020simple, caron2020unsupervised}, to extract view-consistent representations. According to the conclusion of previous methods~\cite{tian2020contrastive, DBLP:conf/iclr/Federici0FKA20}, we found that maximizing the transformation invariance in contrastive learning methods is equivalent to maximizing the intra-view consistency. Furthermore, to improve the generalizability of consistent representations, we assume that a good view-consistent representation needs to satisfy two conditionals: i) transformation invariance and ii) clustering consistency. This part is depicted in section~\ref{sec:mine-consist}. After that, we extract comprehensive representations using the Conditional VAE (CVAE)~\cite{sohn2015learning} and then minimize the upper bound of mutual information between the comprehensive representation and the consistent representation for disentanglement. The benefit of this strategy is that it reduces the complexity of disentangling by reducing the number of unknown variables. This part will be discussed in section~\ref{sec:mine-spec}. We present the framework of the proposed method in Figure~\ref{pic:framework}.


\subsection{Mining Consistency}
\label{sec:mine-consist}
As previously mentioned, we assume that view-consistent information can be extracted through the transformation invariance and clustering consistency. Transformation invariance means that two views of an object, $X^{(1)}$ and $X^{(2)}$, should maintain the same semantic information, regardless of any transformations applied to them. For example, even after applying color jitter, the semantics of $X^{(1)}$ and $X^{(2)}$ should remain unchanged. On the other hand, clustering consistency implies that the clustering prototypes obtained from two views should remain the same after applying the same clustering mapping function. For instance, when processing different views of a dog's front and side, the clustering algorithm should assign both views to the same cluster. Next, we will examine how to extract them from multiple views data. 

In contrastive learning~\cite{chen2020simple, he2020momentum}, the goal is to learn transformation invariance (or augmentation invariance) by reducing the semantic distance between two distinct augmentations of an image. Meanwhile, in ~\cite{tian2020contrastive}, researchers have shown that the intra-view consistent information can be learned by using two different views of an object in multi-view scenarios. Thus, we design the strategy of maximizing multi-view transformation invariance. In essence, we apply an augmentation strategy, similar to \cite{chen2020simple}, to each view. This approach yields a significant advantage in that we can obtain $2V\times$ positive and negative sample pairs. Then, we develop a multi-view contrastive learning loss, which is presented in the following:
\begin{equation}
\label{eq:mv-ins-loss}
\mathcal{L}_{ins} = \frac{1}{VB}\sum_{1 \leq i < j < \leq V} \sum_{k=1}^{2BV}	-\log \frac{h(\boldsymbol{z^{(i)}_k}, \boldsymbol{z^{(j)}_k})}{\sum_{m=1}^{2BV} \mathbbm{1}_{[m \neq k]}h(\boldsymbol{z^{(i)}_k}, \boldsymbol{z^{(j)}_m})}
\end{equation}
where $h(\boldsymbol{a}, \boldsymbol{b}) = \exp ({\rm sim}(\boldsymbol{a}, \boldsymbol{b}) / \tau)$, ${\rm sim}(\boldsymbol{a}, \boldsymbol{b}) = \frac{\boldsymbol{a}^{T}\boldsymbol{b}}{\|\boldsymbol{a}\| \| \boldsymbol{b}\| }$, $B$ is the size of a minibatch, $\mathbbm{1}_{[m \neq k]} \in \{0, 1 \}$ implies an indicator function evaluating to $1$ if $m \neq k$, and $\boldsymbol{z}^{(i)}$ is the contrastive vector of $i$-th view data $x^{i}$ obtained from $\boldsymbol{z}^{(i)} = f(E_{c}(x^{(i)}))$, $f(\cdot)$ is the contrastive head consists of MLP and $E_c(\cdot)$ denotes the consistent encoder.

To further mine the view-consistent information, we argue that different views of an object must have consistent clustering prototypes. Inspired by~\cite{van2020scan}, we encourage different views of an object and $K$ nearest neighbors that are close in embedding space and can be classified into the same cluster. We employ a clustering head, denoted as $g(\cdot)$, to classify each sample in $\mathcal{X}$, which terminates in a softmax function to execute a soft assignment over clusters $\mathcal{C}=\{1, \cdots, C\}$. The probability of assigning $X^{(i)}$ to cluster $c$ is denoted as $g^{c}(X^{(i)})$. Therefore, we define the loss function of multi-view clustering consistency as the following:

\begin{equation}
\label{eq:mv-cls-loss}
\begin{split}
	\mathcal{L}_{clu} = - \frac{1}{VB} \sum_{1 \leq i < j < \leq V} \sum_{k=1}^{B} & \log \langle g(x_k^{(i)}), g(x_k^{(j)}) \rangle \\
	& +\lambda_{clu} \sum_{c \in \mathcal{C}} g^{\prime c} \log g^{\prime c} \\
	& {\rm with \ } g^{\prime c} = \frac{1}{B} \sum_{k=1}^{B} g^{c}(x^{(i)}_{k})
\end{split}
\end{equation}
where $\lambda_{clu}$ denotes the entropy weight, $\langle \cdot \rangle $ is the dot product operator. In Eq. \eqref{eq:mv-cls-loss}, the first term encourages $g(\cdot)$ to make consistent cluster for a sample $x_k^{(i)}$ and its multi-view neighboring sample $x_k^{(j)}$. To avoid $g(\cdot)$ obtaining trivial solutions, the second term is used to spread the clustering results uniformly.

In practice, we have discovered that pre-training the network using Eq. \eqref{eq:mv-ins-loss}, followed by mining the clustering consistency information using Eq. \eqref{eq:mv-cls-loss}, produces superior results compared to training them jointly.

\subsection{Mining Specificity}
\label{sec:mine-spec}

The goal of disentangling the multi-view representations is to separate the consistent representation $S$ and view-specific representations $P^{(v)}$ from the multi-view comprehensive representation $H$. In the unsupervised setting, it is difficult for the model to distinguish which part is the consistency, and which part is the specificity. We have reconsidered the representation disentangling from the information theory perspective. Disentangling specificity from the comprehensive representation is equivalent to minimizing the upper bound of the mutual information between the view-consistent representation and comprehensive representations. Intuitively, we assume that the information of the comprehensive representation equals the sum of the consistent and view-specific information. Like solving ternary equations, solving view-specific representations will be easy when consistent and comprehensive representations are known. Therefore, we can determine the $i$-th view-specific representations $P^{(i)}$ by minimize the following objective:

\begin{equation}
\label{eq:mi-upper-bound}
\min \ I(S, H^{(i)}) + \epsilon
\end{equation}
where $\epsilon$ is the noise contained within the view, and it is a constant. In the complete view setting, we consider $\epsilon$ negligible. Then, we have:

\begin{equation}
\label{eq:mi-ub}
\begin{split}
	I(S, H^{(i)}) & = \mathbb{E}_{q(S, H^{(i)})} \log \frac{q_i(H^{(i)} | S)}{q(H^{(i)})} \\
	   & = \mathbb{E}_{q(S, H^{(i)})} \log \frac{q_i(H^{(i)} | S)}{r(H^{(i)})} \frac{r(H^{(i)})}{q(H^{(i)})} \\
	   & = \mathbb{E}_{q(S, H^{(i)})} \log \frac{q_i(H^{(i)} | S)}{r(H^{(i)})} - \mathbf{KL} [r(H^{(i)}) \| q(H^{(i)}) ] \\
	   & \leq \mathbb{E}_{q(S, H^{(i)})} \log \frac{q_i(H^{(i)} | S)}{r(H^{(i)})} \\
\end{split}
\end{equation}
where $q_i(\cdot)$ denotes the $i$-th view-specific marginal distribution, $q(a ; b)$ implies the joint distribution of random variables $a$ and $b$, and $r(\cdot)$ is mixed Gaussian distributions. In Eq. \eqref{eq:mi-ub}, we can clearly see that the upper bound depends on the approximation of the marginal distribution $r(H^{(i)})$ to prior $q(H^{(i)})$. Since $\mathbf{KL} [r(H^{(i)}) \| q(H^{(i)}) ] \geq 0$, we just need to optimize the first term. Therefore, we simplify the upper bound as the following:
\begin{equation}
\label{eq:obj-ub}
	\min I(S, H^{(i)}) \equiv \min \mathcal{L}_{dis} =  \mathbb{E}_{q(S, H^{(i)})} \log \frac{q_i(H^{(i)} | S)}{r(H^{(i)})}
\end{equation}
Theoretically, as long as the consistent information is accurate enough, Eq. \eqref{eq:obj-ub} can approach the optimal solution indefinitely. This means that the quality of disentangling improves with the improvement of consistent information.

Next, we adopt the architecture of CVAE to extract comprehensive representations $H^{(i)}$. There are two advantages to this approach: i) it allows for fitting disentangled view-specific representations using a mixture of Gaussian distributions, and ii) by combining the category information output by the consistent encoder, the interpretability of the representations can be improved. We extend the CVAE to multi-view scenarios, and its loss function is as follows:

\begin{equation}
\label{eq:cave-loss}
\begin{split}
	\mathcal{L}_{cave} = - \sum_{i=1}^{V} \ & \mathbb{E}_{\mathbf{z} \sim q(\mathbf{z} | X^{(i)}, S)}[\log \Phi (X^{(i)} | \mathbf{z}, S) ] \\
	& + \mathbf{KL} [q(\mathbf{z} | X^{(i)}, S) \| \Phi(\mathbf{z})) ]
\end{split}	
\end{equation}
where $\mathbf{z} = [P^{(i)};S]$, $[\cdot]$ denotes the concatenating operation, and $P^{(i)} \sim \mathcal{N}(\mathbf{0}, \mathbf{1})$. Therefore, we can formulate the joint loss function of the second stage:
\begin{equation}
	\mathcal{L}_{spc} = \mathcal{L}_{cave} + \lambda_{dis} \mathcal{L}_{dis}.
\end{equation}


\section{Experiments}
\label{sec:experiments}

\subsection{Dataset}
\label{sec:dataset}
We evaluate the proposed method and other competitive methods using four multi-view datasets. There are: (a) Edge-MNIST \cite{liu2016coupled}, which is a well-known benchmark dataset consisting of 70,000 grayscale digit images (0-9) with $28 \times 28$ pixels. The views contain the original digits and the edge-detected version, respectively; (b) Edge-FMNIST \cite{xiao2017fashion}, which is a fashion dataset consisting of $28 \times 28$ grayscale images of clothing items. We synthesize the second view by running the same edge detector used to create Edge-MNIST; (c) COIL-20 \cite{nene1996columbia}, which depicts from different angles containing grayscale images of 20 items. We create a two-view dataset by randomly grouping the images for an item into two groups; (d) MVC-10 \cite{liu2016mvc}, which is a multi-angle clothing dataset consisting of 161,260 left-, right-, back-, and front-view with 10 categories. In our experiments, we use any three views to build the multi-view dataset. We report the dataset description in Table ~\ref{dataset_description}.

\begin{table}[t]
\renewcommand{\arraystretch}{1.3}
\caption{Dataset Description}
\label{dataset_description}       
\begin{center}
\begin{tabular}{ccccc}
\hline\noalign{\smallskip}
Dataset & \#samples & \#view & \#class & \#shape \\
\noalign{\smallskip}\hline\noalign{\smallskip}
Edge-MNIST & 70,000 & 2 & 10 & (1 $\times$ 28 $\times$ 28) \\
Edge-FMNIST & 70,000 & 2 & 10 &  (1 $\times$ 28 $\times$ 28) \\
COIL-20 & 1,440 & 2 & 20 & (1 $\times$ 128 $\times$ 128) \\
MVC-10  & 161,260 & 3 & 10 & (3 $\times$ 224 $\times$ 224) \\
\noalign{\smallskip}\hline
\end{tabular}
\end{center}
\end{table}

\subsection{Baseline and Metrics}\label{sec:baseline-models}
We compare the proposed method and the following 12 baseline clustering and classification methods, which are categorized into three types: \textbf{(a) Single-view methods:} K-means (KM) for clustering, and Support Vector Machine (SVM) for classification. Note that KM$_{cat}$ denotes concatenating all view-specific representations. $\beta$-VAE~\cite{higgins2017beta} is a VAE-based method, which can obtain disentangled representation in the single-view scenario. In our settings, we select the best view as the $\beta$-VAE's input; \textbf{(b) Multi-view methods:} SCAN~\cite{van2020scan} and SimCLR~\cite{chen2020simple} are self-supervised methods, and we use them to extract the consistent representation. SiMVC and CoMVC ~\cite{trosten2021reconsidering} are two contrastive learning-based multi-view clustering methods. EAMC~\cite{zhou2020eamc} is an adversarial multi-view clustering method. CMC~\cite{tian2020contrastive} is a contrastive multi-view representation learning method. MORI-RAN~\cite{DBLP:conf/icdm/KeZY22} is a contrastive fusion-based multi-view representation learning method; \textbf{(c) Multi-view disentangled methods:} Multi-VAE~\cite{xu2021multi} is a VAE-based multi-view disentangled representation learning method. MIB~\cite{DBLP:conf/iclr/Federici0FKA20} is an information bottleneck-based multi-view representation learning method. Note that MIB is limited to two views; for datasets with more than two views, we select the best two views as its input.

\subsection{Implementation Details}
We implement the proposed method and other non-linear comparison methods on the PyTorch 1.10 \cite{paszke2019pytorch} platform, running on Ubuntu 18.04 LTS utilizing an NVIDIA A100 tensor core Graphics Processing Units (GPUs) with 40 GB memory size. For simplicity, we use ResNet~\cite{he2016deep} as the consistent encoder, where ResNet-18 is used for the Edge-MNIST and Edge-FMNIST datasets, and ResNet-34 is used for the COIL-20 and MVC-10 datasets. We set the dimensionality of view-specific encoders to $I-{\rm Conv}_{32}^{4}-{\rm Conv}_{64}^{4}-{\rm Conv}_{128}^{4}-{\rm Conv}_{256}^{4}-O$ for all experiments, where $I$ and $O$ indicate the dimension of the data's input and the encoder's output, respectively. It means that convolution kernel sizes are $4-4-4$, channels are $32-64-128-256$, the stride is set as $2$, and the dimensionality of embedding is $256$. The decoders are symmetric with the encoders. For all $\mu^{v}$ and $\sigma^{v}$, are set as $10$-dimensional. We pre-train the consistent encoder using a similar way in the ~\cite{chen2020simple} and ~\cite{van2020scan}. For the second stage of our method, we use Adam with default parameters and an initial learning rate of $0.0005$ for training view-specific encoders and decoders for 150 epochs. For the comparing methods, we use their release codes with the settings recommended by the authors. For evaluation, we extract all latent representations, then feed them into K-means and SVM, and report their results, respectively. To eliminate the randomness, we run our method and other methods 10 times, and report their average and standard deviation values in terms of all evaluation metrics.

\begin{table*}[t]
\renewcommand{\arraystretch}{1.3}
\setlength\tabcolsep{2pt}
\caption{Clustering results on four datasets, where “–” denotes the dataset cannot handle such scenarios. $^\dagger$ indicates that we use Eq. \eqref{eq:mv-ins-loss} and \eqref{eq:mv-cls-loss} to enable it to handle multi-view data. The best and the second best values are highlighted in {\color{red}red} and {\color{blue}blue}, respectively. All results are reproduced using our implemented code.}
\label{tab:clustering-result}       
\begin{center}
\scalebox{0.82}{
\begin{tabular}{ccccccccccccc}
\hline\noalign{\smallskip}
 & \multicolumn{3}{c}{Edge-MNIST} & \multicolumn{3}{c}{Edge-FMNIST} & \multicolumn{3}{c}{COIL-20} & \multicolumn{3}{c}{MVC-10}  \\
\noalign{\smallskip} Method & ACC$_{clu}$ & NMI & ARI & ACC$_{clu}$ & NMI & ARI & ACC$_{clu}$ & NMI & ARI & ACC$_{clu}$ & NMI & ARI  \\
\noalign{\smallskip}\hline\noalign{\smallskip}
KM$_{cat}$ & 38.27$\pm$1.93 & 32.87$\pm$1.69 & 20.07$\pm$1.30 & 21.82$\pm$1.08 & 21.71$\pm$1.25 & 14.36$\pm$1.63 & 36.25$\pm$3.06 & 50.38$\pm$1.55 & 22.28$\pm$2.95 & - & - & -  \\
$\beta$-VAE (NIPS'18)~\cite{higgins2017beta} & 57.88$\pm$0.42 & 52.77$\pm$0.02 & 48.17$\pm$0.01 & 40.87$\pm$0.11 & 39.48$\pm$0.32 & 39.42$\pm$0.13 & 18.52$\pm$0.53 & 58.73$\pm$0.60 & 34.12$\pm$0.60 & 34.55$\pm$1.07 & 30.45$\pm$0.86 & 11.03$\pm$1.37  \\
\noalign{\smallskip}\hline\noalign{\smallskip}
SimCLR$^\dagger$ (PRML'20)~\cite{chen2020simple} & 44.69$\pm$0.26 & 36.90$\pm$0.12 & 27.06$\pm$0.31 & 47.48$\pm$0.42 & 47.14$\pm$0.31 & 31.29$\pm$0.21 & 80.15$\pm$3.63 & 92.44$\pm$0.92 & 80.76$\pm$3.10 & 38.64$\pm$1.82 & 41.30$\pm$1.45 & 21.78$\pm$1.59  \\
SCAN$^\dagger$ (ECCV'20)~\cite{van2020scan} & 95.38$\pm$1.26 & \textcolor{blue}{91.44$\pm$1.28} & \textcolor{blue}{90.19$\pm$1.73} & \textcolor{blue}{70.34$\pm$1.95} & \textcolor{blue}{65.82$\pm$1.86} & \textcolor{blue}{57.19$\pm$1.44} & \textcolor{blue}{89.11$\pm$2.30} & 94.57$\pm$3.42 & 88.26$\pm$2.37 & 39.08$\pm$1.53 & \textcolor{blue}{45.14$\pm$1.32} & \textcolor{blue}{31.65$\pm$1.18}  \\
SiMVC (CVPR'21)~\cite{trosten2021reconsidering} & 86.41$\pm$1.15 & 83.17$\pm$0.94 & 82.66$\pm$1.17 & 56.83$\pm$0.27 & 50.12$\pm$0.15 & 43.28$\pm$0.31 & 77.45$\pm$2.59 & 92.03$\pm$3.16 & 73.58$\pm$6.33 & 35.82$\pm$4.18 & 38.61$\pm$3.47 & 20.97$\pm$3.92  \\
CoMVC (CVPR'21)~\cite{trosten2021reconsidering} & \textcolor{blue}{95.42$\pm$1.54} & 90.85$\pm$2.38 & 89.73$\pm$2.05 & 58.94$\pm$2.69 & 53.27$\pm$2.08 & 49.72$\pm$3.22 & 90.04$\pm$1.20 & \textcolor{blue}{95.19$\pm$1.21} & \textcolor{red}{91.13$\pm$1.38} & 39.15$\pm$2.19 & 43.96$\pm$1.67 & 25.79$\pm$1.52  \\
EAMC (CVPR'20)~\cite{zhou2020eamc} & 66.78$\pm$0.52 & 63.11$\pm$0.34 & 59.83$\pm$1.22 & 55.21$\pm$1.60 & 62.57$\pm$1.48 & 48.51$\pm$1.88 & 67.28$\pm$3.59 & 75.83$\pm$4.21 & 68.73$\pm$6.30 & \textcolor{blue}{41.38$\pm$1.08} & 42.53$\pm$0.66 & 29.46$\pm$1.35  \\
CMC (ECCV'20)~\cite{tian2020contrastive} & 80.76$\pm$1.12 & 78.49$\pm$1.18 & 76.28$\pm$0.78 & 49.56$\pm$2.25 & 45.68$\pm$2.16 & 43.55$\pm$1.47 & 78.32$\pm$1.38 & 91.20$\pm$1.33 & 63.40$\pm$1.32 & 25.17$\pm$2.54 & 19.34$\pm$2.93 & 15.10$\pm$1.09  \\
MORI-RAM (ICDM'22)~\cite{DBLP:conf/icdm/KeZY22} & 82.13$\pm$4.11 & 78.25$\pm$1.68 & 75.14$\pm$3.52 & 58.33$\pm$4.48 & 55.19$\pm$2.70 & 40.86$\pm$2.23 & 73.58$\pm$0.71 & 84.72$\pm$1.36 & 71.95$\pm$1.19 & 24.48$\pm$0.29 & 22.39$\pm$0.49 & 12.29$\pm$1.05  \\
\noalign{\smallskip}\hline\noalign{\smallskip}
Multi-VAE (CVPR'21)~\cite{xu2021multi} & 60.24$\pm$0.83 & 58.37$\pm$0.58 & 44.16$\pm$0.91 & 53.38$\pm$1.14 & 56.56$\pm$1.56 & 41.02$\pm$0.95 & 64.58$\pm$1.58 & 79.59$\pm$1.15 & 54.19$\pm$2.50 & 33.49$\pm$0.13 & 34.72$\pm$0.44 & 17.09$\pm$1.10  \\
MIB (ICLR'20)~\cite{DBLP:conf/iclr/Federici0FKA20} & 53.65$\pm$5.33 & 48.16$\pm$7.41 & 36.07$\pm$3.66 & 54.41$\pm$4.82 & 53.08$\pm$6.50 & 44.69$\pm$6.77 & 51.67$\pm$5.79 & 83.12$\pm$3.36 & 56.76$\pm$4.89 & - & - & -  \\
\noalign{\smallskip}\hline\noalign{\smallskip}
Ours & \textcolor{red}{97.71$\pm$1.24} & \textcolor{red}{95.82$\pm$2.07} & \textcolor{red}{95.08$\pm$1.28} & \textcolor{red}{73.06$\pm$1.47} & \textcolor{red}{70.39$\pm$0.28} & \textcolor{red}{60.44$\pm$1.44} & \textcolor{red}{90.64$\pm$2.24} & \textcolor{red}{97.36$\pm$1.07} & \textcolor{blue}{90.68$\pm$2.33} & \textcolor{red}{48.91$\pm$0.56} & \textcolor{red}{47.67$\pm$0.73} & \textcolor{red}{33.65$\pm$0.66}  \\
$\Delta$ SOTA  & \textcolor{forestgreen}{$\uparrow$2.29} & \textcolor{forestgreen}{$\uparrow$4.37} & \textcolor{forestgreen}{$\uparrow$4.89} & \textcolor{forestgreen}{$\uparrow$2.72} & \textcolor{forestgreen}{$\uparrow$4.57} & \textcolor{forestgreen}{$\uparrow$3.25} & \textcolor{forestgreen}{$\uparrow$1.53} & \textcolor{forestgreen}{$\uparrow$2.17} & \textcolor{gray}{$\downarrow$0.45} & \textcolor{forestgreen}{$\uparrow$7.53} & \textcolor{forestgreen}{$\uparrow$2.53} & \textcolor{forestgreen}{$\uparrow$2.00}\\
\noalign{\smallskip}\hline
\end{tabular}}
\end{center}
\end{table*}

\begin{table*}[t]
\renewcommand{\arraystretch}{1.3}
\caption{Classification results on four datasets, where “–” denotes the dataset cannot handle such scenarios. $^\dagger$ indicates that we use Eq. \eqref{eq:mv-ins-loss} and \eqref{eq:mv-cls-loss} to enable it to handle multi-view data. The best and the second best values are highlighted in {\color{red}red} and {\color{blue}blue}, respectively. All results are reproduced using our implemented code.}
\label{tab:classification-result}       
\begin{center}
\begin{tabular}{ccccccccc}
\hline\noalign{\smallskip}
 & \multicolumn{2}{c}{Edge-MNIST} & \multicolumn{2}{c}{Edge-FMNIST} & \multicolumn{2}{c}{COIL-20} & \multicolumn{2}{c}{MVC-10}  \\
\noalign{\smallskip} Method & ACC$_{cls}$    & F-Score & ACC$_{cls}$    & F-Score & ACC$_{cls}$    & F-Score & ACC$_{cls}$    & F-Score  \\
\noalign{\smallskip}\hline\noalign{\smallskip}
SVM$_{cat}$ & 42.89$\pm$0.07    & 41.11$\pm$0.5 & 53.51$\pm$0.21    & 53.58$\pm$0.08 & 10.42$\pm$0.01   & 8.02$\pm$0.01 & -  & -  \\
$\beta$-VAE (NIPS'18)~\cite{higgins2017beta} & 96.11$\pm$0.14   & 96.06$\pm$0.16 & 81.61$\pm$0.03  & \textcolor{blue}{81.50$\pm$0.41} & 95.49$\pm$0.25 &  96.04$\pm$0.03 & 70.63$\pm$2.13  & 57.14$\pm$1.04  \\
\noalign{\smallskip}\hline\noalign{\smallskip}
SimCLR$^\dagger$ (PRML'20)~\cite{chen2020simple} & 97.97$\pm$0.02  & 97.95$\pm$0.03 & 80.09$\pm$0.03   & 80.06$\pm$0.02 & 97.19$\pm$0.11  & 97.00$\pm$0.11 & 71.61$\pm$0.20  & 67.02$\pm$0.16  \\
SCAN$^\dagger$ (ECCV'20)~\cite{van2020scan} & \textcolor{blue}{98.13$\pm$0.04}  & \textcolor{blue}{98.04$\pm$0.10} & \textcolor{blue}{83.62$\pm$0.02} &   80.16$\pm$0.01 & \textcolor{blue}{98.60$\pm$0.01} &  \textcolor{blue}{98.60$\pm$0.01} & \textcolor{blue}{72.84$\pm$0.14} & \textcolor{blue}{71.28$\pm$0.11}  \\
CMC (ECCV'20)~\cite{tian2020contrastive} & 97.53$\pm$0.03 &  97.50$\pm$0.02 & 77.11$\pm$0.14  & 75.78$\pm$0.28 & 97.31$\pm$0.01   & 97.31$\pm$0.01 & 70.19$\pm$0.89  & 69.79$\pm$1.42  \\
MORI-RAM (ICDM'22)~\cite{DBLP:conf/icdm/KeZY22} & 94.76$\pm$0.94  & 94.14$\pm$0.56 & 77.88$\pm$1.12   & 77.16$\pm$0.99 & 88.43$\pm$1.01  & 85.69$\pm$1.33 & 65.26$\pm$0.18 & 61.77$\pm$0.69  \\
MIB (ICLR'20)~\cite{DBLP:conf/iclr/Federici0FKA20} & 90.81$\pm$1.30 & 90.03$\pm$0.69 & 75.33$\pm$0.05 & 73.80$\pm$0.05 & 59.72$\pm$2.29  & 53.99$\pm$2.03 & - &  -  \\
\noalign{\smallskip}\hline\noalign{\smallskip}
Ours & \textcolor{red}{99.41$\pm$0.06}  & \textcolor{red}{99.41$\pm$0.04} & \textcolor{red}{85.19$\pm$0.07} & \textcolor{red}{84.88$\pm$0.07} & \textcolor{red}{99.81$\pm$0.01}  & \textcolor{red}{99.80$\pm$0.01} & \textcolor{red}{79.07$\pm$0.09} & \textcolor{red}{79.04$\pm$0.22}  \\
$\Delta$ SOTA  & \textcolor{forestgreen}{$\uparrow$1.28} &  \textcolor{forestgreen}{$\uparrow$1.37} & \textcolor{forestgreen}{$\uparrow$1.57}   & \textcolor{forestgreen}{$\uparrow$3.38} & \textcolor{forestgreen}{$\uparrow$1.21} &  \textcolor{forestgreen}{$\uparrow$1.20} & \textcolor{forestgreen}{$\uparrow$6.20} &  \textcolor{forestgreen}{$\uparrow$7.76}\\
\noalign{\smallskip}\hline
\end{tabular}
\end{center}
\end{table*}

\subsection{Evaluation Metrics}
In order to evaluate clustering performance, three standard evaluation metrics are used: clustering ACCuracy (ACC$_{clu}$), Normalized Mutual Information (NMI), and Adjusted Rand Index (ARI). Readers seeking further details on these metrics are referred to \cite{kumar2011co}. It is important to note that the validation process of clustering methods is limited to cases where ground truth labels are available. For classification, ACC$_{cls}$ and F-Score are utilized as evaluation metrics. In all cases, a higher value indicates better performance.


\subsection{Comparison Results and Analysis}

We evaluate our method with 12 state-of-the-art multi-view methods in terms of clustering and classification performance on four datasets, as shown in Table~\ref{tab:clustering-result} and Table~\ref{tab:classification-result}, where the following observations are obtained: 

Our method outperforms other compared methods on all metrics of clustering and classification tasks. Especially, we achieve significant improvements over the second-best method on Edge-MNIST, Edge-FMNIST, and MVC-10 datasets. For instance, our method achieves $2.29\%$ $(97.71\%-95.42\%)$ and $7.53\%$ $(48.91\%-41.38\%)$ higher clustering accuracy than the second-best method on Edge-MNIST and MVC-10 datasets, respectively. 

In comparison with single-view methods (KM$_{cat}$ and $\beta$-VAE), we find that simply concatenating all comprehensive representations of different views does not significantly improve the downstream performance. Furthermore, the multi-view disentangling method (Multi-VAE, MIB, and our method) significantly outperforms the single-view disentangling method. Therefore, we believe that disentangling the integrated representation is beneficial for improving the performance of downstream tasks. The essential reason is that disentangling can extract some task-independent information. 

Compared with other contrastive learning-based methods, we find that using only consistent representations can achieve satisfactory results in clustering and classification tasks. At the same time, we also find that adding some specificity information appropriately can further improve the performance of downstream tasks. For example, both our method and Multi-VAE~\cite{xu2021multi} concatenate the consistent representation and view-specific representations into one. We believe that this paradigm helps the model to process the information for the required part of the downstream tasks. 

In the comparison results between our method and the end-to-end disentangling method (Multi-VAE~\cite{xu2021multi} and MIB~\cite{DBLP:conf/iclr/Federici0FKA20}), we find that as the complexity of data (quantity and dimension) increases, it becomes increasingly difficult to disentangle the consistency and specificity simultaneously from the comprehensive representation. In contrast, our method becomes more prominent in reducing the complexity of disentangling. These results confirm our observation that the boundary between multi-view consistency and specificity becomes unclear in the unsupervised setting. Therefore, obtaining one part of the information can improve the quality and reduce the difficulty of disentangling.

\begin{figure}[t!]
\centering
\includegraphics[width=3.2in]{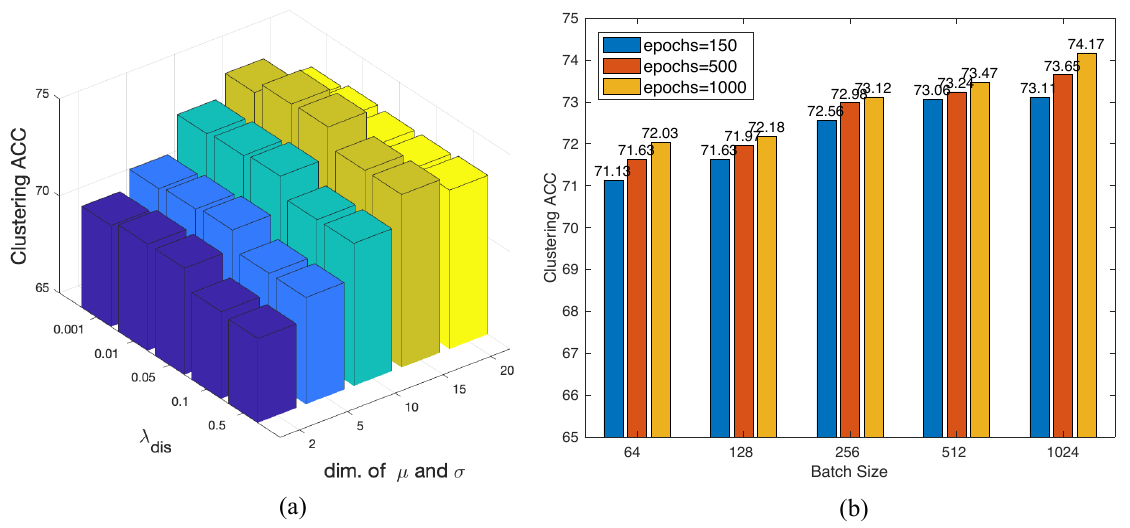}
\caption{(a) Parameter sensitivity analysis of our method's performance to changes in the dimensionality of $\mu$ and $\sigma$ and the hyperparameter $\lambda_{dis}$. (b) The effect of varying batch size and number of training epochs on clustering performance.}
\label{pic:parameters_ana}
\end{figure}

\subsection{Ablation Study}
We conducted a comprehensive ablation study on Edge-FMNIST, including the proposed method without pretext task $\mathcal{L}_{ins}$, the proposed method without the pseudo-label prediction $\mathcal{L}_{clu}$, and the proposed method without disentangling module $\mathcal{L}_{spc}$. We compared these components with our complete method, and the results are shown in Table~\ref{tab:ab_module}. One intuitive result is that using the disentangling module can improve the performance of the model, with $2.72\%$ $(73.06\%-70.34\%)$ increase in terms of $ACC_{clu}$ metric. Additionally, we find that without $\mathcal{L}_{ins}$, the pseudo-label prediction cannot work independently, leading to poor results. At the same time, when only the disentangling module is used, its performance is relatively poor. This suggests that our method heavily relies on the quality of consistency, and the higher the quality of the consistent representation, the better the quality of disentangling.

\begin{table}[t]
\renewcommand{\arraystretch}{1.3}
\caption{Components analysis on the E-FMNIST dataset.}
\label{tab:ab_module}       
\begin{center}
\scalebox{0.95}{\begin{tabular}{cccccccc}
\hline\noalign{\smallskip}
$\mathcal{L}_{ins}$ & $\mathcal{L}_{clu}$ & $\mathcal{L}_{spc}$ & ACC$_{clu}$ & NMI & ARI & ACC$_{cls}$ & F-score\\
\noalign{\smallskip}\hline\noalign{\smallskip} 
$\checkmark$ & $\checkmark$ &  $\checkmark$ & 73.06 & 70.39 & 60.44 & 85.19 &  84.88 \\
\noalign{\smallskip}\hline\noalign{\smallskip} 
$\checkmark$ &  & $\checkmark$ & 49.34 & 50.11 & 42.40 & 80.46 & 78.39 \\
& $\checkmark$ &  $\checkmark$ & 31.77 & 27.19 & 24.38 & 51.06 & 49.37 \\
 & & $\checkmark$ & 37.15 & 30.22 & 27.43 & 63.84 & 62.10 \\
\noalign{\smallskip}\hline\noalign{\smallskip} 
 $\checkmark$ &  &  & 47.48 & 47.14 & 31.29 & 80.09 & 80.06 \\
 & $\checkmark$ &   & 12.81 & 9.57 & 2.05 & 10.05  & 10.05 \\
 $\checkmark$ & $\checkmark$ &  & 70.34 & 65.82 & 57.19 & 83.62 & 80.16 \\
\noalign{\smallskip}\hline
\end{tabular}}
\end{center}
\end{table}

\subsection{Parameter Analysis}

We conducted a hyperparameter analysis of the proposed method on the Edge-FMNIST dataset, including the dimensionality of $\mu$ and $\sigma$, $\lambda_{dis}$, batch size, and training epochs, as shown in Figure~\ref{pic:parameters_ana}. According to the results in Figure~\ref{pic:parameters_ana}(a), we find that the dimensionality of $\mu$ and $\sigma$ range from 10 to 15, and $\lambda_{dis}$ range from 0.01 to 0.05 can achieve better results. the dimensionality of $\mu$ and $\sigma$ too low will lead to insufficient representation, while too high will produce redundant representation. $\lambda_{dis}$ is the penalty coefficient of the disentanglement loss $\mathcal{L}_{dis}$, and if it is too low, disentangling may not be sufficient, while if it is too high, the model may obtain trivial solutions. On the other hand, as shown in Figure~\ref{pic:parameters_ana}(b), with the increase of batch size and epochs, the performance of the proposed method will also increase. We believe that increasing training time will be more beneficial when the batch size is less than 512.

\begin{figure*}[t]
\centering
\begin{overpic}[scale=.68, tics=3]{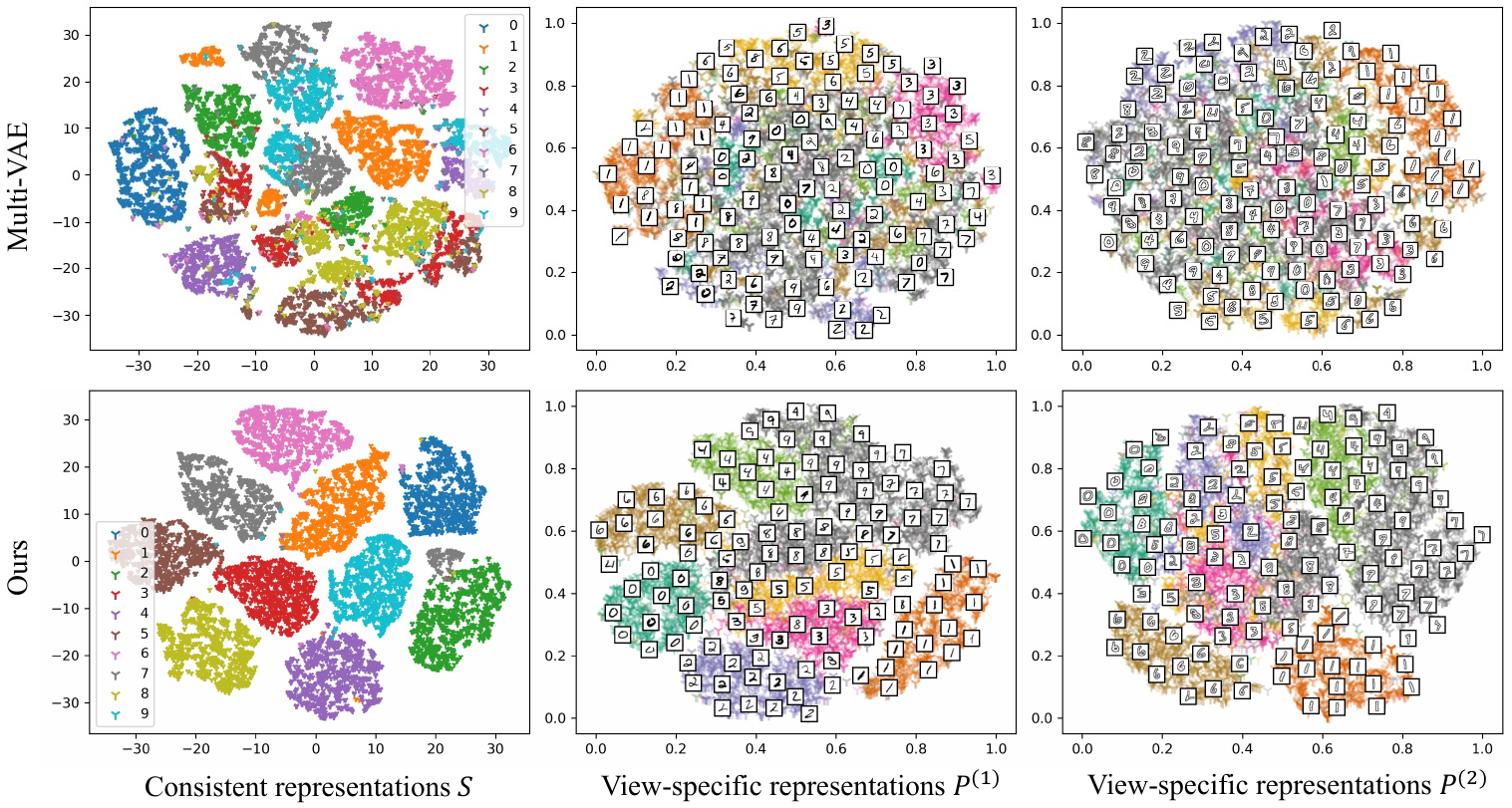}
\put(0.5, 45.2){\rotatebox{90}{~\cite{xu2021multi}}}
\end{overpic}

\caption{Visualization of representations using t-SNE~\cite{van2008visualizing} on the Edge-MNIST dataset. The top row displays the representations of Multi-VAE~\cite{xu2021multi}, while the bottom row shows the representations obtained by our proposed method. The leftmost column corresponds to the consistent representation $S$, while the subsequent two columns display the view-specific representations $P^{(1)}$ and $P^{(2)}$ of two different views, respectively.}
\label{pic:visualization}
\end{figure*}

\begin{figure}[t]
\centering
\includegraphics[width=3.35in]{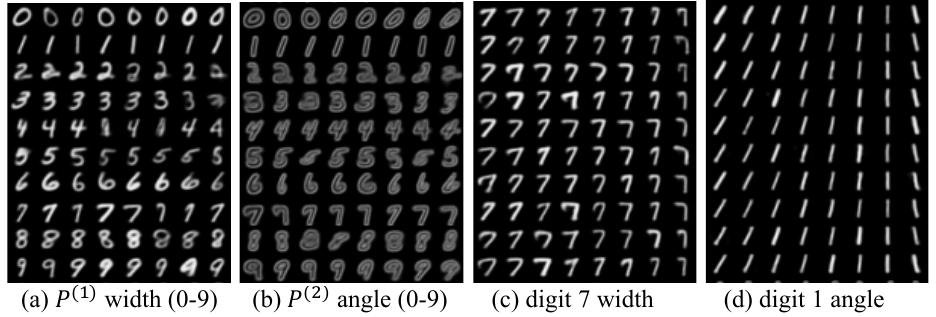}
\caption{Visualization of disentangled representations: (a) Width variation of view-specific representation $P^{(1)}$ for different digits; (b) Angle variation of view-specific representation $P^{(2)}$ for different digits; (c) Width variation of specific digit ``7'' in both views; (d) Angle variation of specific digit ``1'' in both views.}
\label{pic:disentangled}
\end{figure}

\subsection{Visualization}
We visualize all the representations of the proposed method on the Edge-MNIST dataset in the presented results, as shown in Figure~\ref{pic:visualization} and Figure~\ref{pic:disentangled}. In Figure~\ref{pic:visualization}, we use t-SNE~\cite{van2008visualizing} to visualize the consistent representation of Multi-VAE~\cite{xu2021multi} and our method, and view-specific representations. We can see that the consistent representation extracted by our method is more compact than Multi-VAE~\cite{xu2021multi}. Furthermore, in the view-specific space, we find that the specificity extracted by our method can also be divided into specific attribute regions. For example, in the view-specific representation $P^{(1)}$, we can see that there is a significant angle change for the digit ``1'' when the value of the x-axis changes from $0.6$ to $1.0$. In addition, we show the change of digit width and angle in Figure~\ref{pic:disentangled}. Thanks to our disentanglement method, we can generate data with specific attributes and specific categories, such as digits ``1'' with different angle variations and digits ``7'' with different widths. The above results indicate that our disentanglement method can make multi-view representations have good interpretability and compactness.


\section{Conclusion}
\label{sec:conclusion}

In summary, we propose a novel two-stage disentanglement method that mines multi-view consistency by maximizing the transformation invariance and clustering consistency, and mines specificity by minimizing the mutual information between the consistent and comprehensive representations. Our method achieved superior clustering and classification performance on four datasets, and the ablation studies demonstrated the effectiveness of our disentangling module in enhancing the expressive power of concatenated representations. Moreover, the interpretability and compactness of both consistency and specificity obtained by our method were demonstrated through visualization results. In future work, we plan to extend our method to the incomplete-view scenario, as we have observed that our method can effectively generate specific views and help predict and restore missing views in such scenarios.

\section*{Acknowledgment}
This work was supported by the Fundamental Research Funds for the Beijing Jiaotong University (No. 2021JBWZB002 \& No. 2023YJS113); Guangdong Natural Science Funds for Distinguished Young Scholar (No. 2023B1515020097); and Singapore Ministry of Education Academic Research Fund Tier 1 (MSS23C002).

\bibliographystyle{ACM-Reference-Format}
\bibliography{paper.bib}

\end{sloppypar}
\end{document}